\documentclass[sigconf]{acmart}

\usepackage{ulem}
\usepackage{xspace}
\usepackage[]{hyperref}
\usepackage[table,xcdraw]{xcolor}
\usepackage{amsmath,amsfonts}

\usepackage{amssymb}
\usepackage{algorithm}
\usepackage{algorithmic}
\usepackage{graphicx}
\usepackage{textcomp}

\usepackage{color}
\usepackage{stfloats}
\usepackage{url}
\usepackage{verbatim}
\usepackage{listings}
\usepackage{xspace}
\usepackage{booktabs}
\usepackage{multirow}
\usepackage{amssymb}
\usepackage{booktabs}
\usepackage{float}
\usepackage{pifont}
\usepackage{ulem}

\newcommand{\we}{\textit{SpeContext}\xspace}

\begin{document}

\title[]{\we: Enabling Efficient Long-context Reasoning with Speculative Context Sparsity in LLMs}


\author{Jiaming Xu}

\email{jiamingxu@sjtu.edu.cn}
\orcid{0009-0000-7000-6537}
\affiliation{%
  \institution{Shanghai Jiao Tong University; SII}
  \city{Shanghai}
  \country{China}
}

\author{Jiayi Pan}
\email{pan_jiayi@sjtu.edu.cn}
\orcid{0009-0007-0468-4592}
\affiliation{%
  \institution{Shanghai Jiao Tong University; Infinigence-AI}
  \city{Shanghai}
  \country{China}
}

\author{Hanzhen Wang}
\email{alex-wang@sjtu.edu.cn}
\orcid{0009-0000-1335-2855}
\affiliation{%
  \institution{Shanghai Jiao Tong University}
  \city{Shanghai}
  \country{China}
}

\author{Yongkang Zhou}
\email{zeenny.willians@sjtu.edu.cn}
\orcid{0009-0008-7732-6347}
\affiliation{%
  \institution{Shanghai Jiao Tong University; SII}
  \city{Shanghai}
  \country{China}
}

\author{Jiancai Ye}
\email{yejiancai@sjtu.edu.cn}
\orcid{0009-0007-6189-3251}
\affiliation{%
  \institution{Shanghai Jiao Tong University}
  \city{Shanghai}
  \country{China}
}

\author{Yu Wang}
\email{yu-wang@tsinghua.edu.cn}
\orcid{}
\affiliation{%
  \institution{Tsinghua University}
  \city{Beijing}
  \country{China}
}

\author{Guohao Dai}
\authornote{Corresponding Author}
\email{daiguohao@sjtu.edu.cn}
\orcid{0000-0003-0849-3252}
\affiliation{%
  \institution{Shanghai Jiao Tong University; Infinigence-AI; SII}
  \city{Shanghai}
  \country{China}
}

\begin{abstract}

As test-time scaling in large language model(LLM) reasoning has been proven effective in enhancing the model performance through step-by-step generation, this long-context generation incurs substantial Key-Value(KV) cache, posing a critical bottleneck for practical applications deployment(\textit{e.g.}, Agents). While recent KV cache optimizations perform well in the long-context input scenario, the following problems remain unsolved if directly applied to long-context reasoning.
\textbf{(1) Time-consuming layer-wise retrieval operation}. The retrieval operation, which selects the important KV pairs in each layer, brings the synchronization overhead that scales with model depth due to the data dependency, resulting in up to $60\%$ latency overhead.
\textbf{(2) Complete retention of the newly generated KV cache}. Existing works designed for long-context input choose to retain the KV pair of newly generated tokens to avoid repeated, time-consuming processing on the KV cache, rendering them ineffective in long-context reasoning.  
\textbf{(3) Performance degradation with a tiny increase in sequence length}. Existing offloading strategies determined before inference cannot adapt to the increasing sequence length, resulting in $>80\%$ performance degradation with a tiny increase in sequence length.

In this paper, we point out that the objective of the retrieval algorithms is to align with the LLM, which is similar to the objective of knowledge distillation in LLMs. We analyze the similarity in information focus between the distilled language model(DLM) and the original LLM from the perspective of information theory, and thus propose a \textbf{novel paradigm that leverages a DLM as the retrieval algorithm}. Based on the insight, we present \we, an algorithm and system co-design for long-context reasoning. 
\textbf{(1) At the algorithm level,} \we 
proposes \textbf{lightweight retrieval head} based on the head-level attention weights of DLM, achieving $>90\%$ parameters reduction by pruning the redundancy. 
\textbf{(2) At the system level, } \we designs an \textbf{asynchronous prefetch dataflow via the elastic loading strategy}, effectively overlapping KV cache retrieval with the LLM computation.
\textbf{(3) At the compilation level,} \we constructs the theoretical memory model and implements an \textbf{adaptive memory management system} to achieve acceleration by maximizing GPU memory utilization.
We deploy and evaluate \we in two resource-constrained environments, cloud and edge. 
Extensive experiments show that, compared with the Huggingface framework, \we achieves up to $\textbf{24.89}\times$ throughput improvement in cloud and $\textbf{10.06}\times$ speedup in edge with negligible accuracy loss, pushing the Pareto frontier of accuracy and throughput.

\end{abstract}

\maketitle

\begin{figure}[!t]
    \centering
    \includegraphics[width=0.48\textwidth]{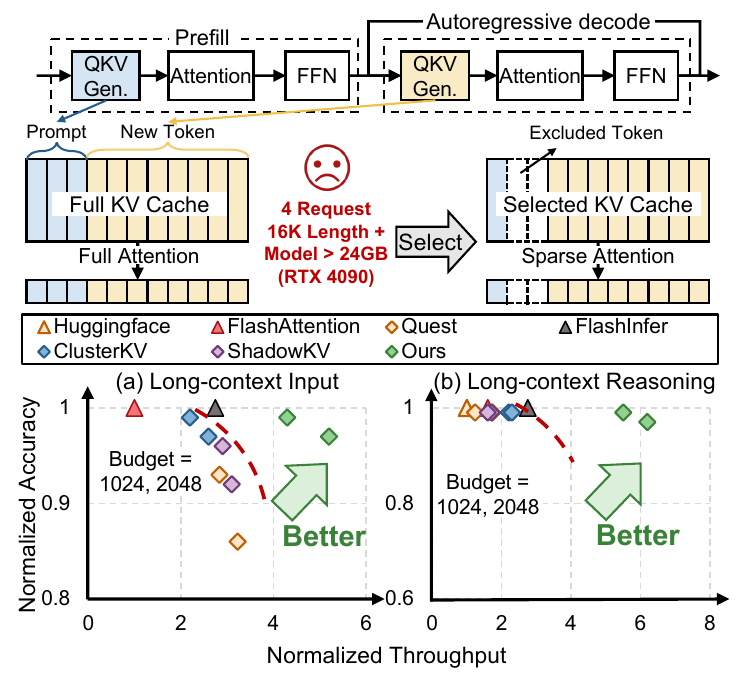}
    \vspace{-20pt}
    \caption{(a)(b) Pareto frontiers on KV cache selection in long-context input and reasoning scenarios.}
    \vspace{-20pt}
    \label{fig:pareto}
\end{figure}

\section{Introduction} \label{sec:intro}

\begin{figure*}[!t]
    \centering
    
    \includegraphics[width=0.98\textwidth]{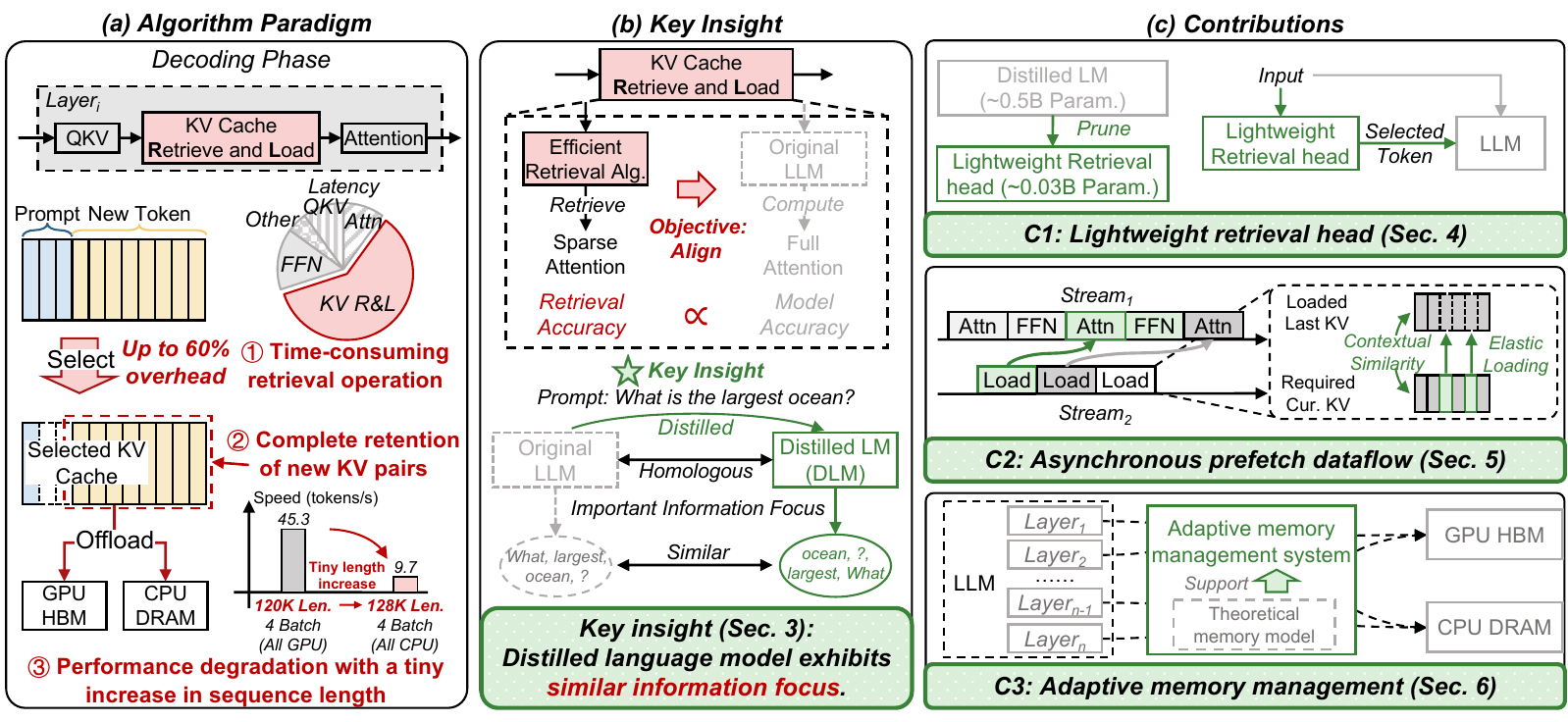}
    \vspace{-10pt}
    \caption{Overview of \we. (a) Three challenges in existing algorithm paradigm in the long-context reasoning scenario. (b) Key Insight: Distilled language model exhibits similar information focus. (c) \uline{C}ontributions from Section~\ref{sec:T1} to Section~\ref{sec:T3}}
    \vspace{-6pt}
    \label{fig:overview}
\end{figure*}

Generative large language models(LLMs) mark a significant advancement in the pursuit of Artificial General Intelligence(AGI). Their successful application across various domains has greatly contributed to the rapid advancement of numerous downstream tasks (\textit{e.g.}, pharmaceutical~\cite{pal2023chatgpt}, finance~\cite{li2023large}, and ecology~\cite{morera2024foundation}), and their remarkable capabilities have attracted widespread attention, spurring the development of a new wave of LLM-based software applications (\textit{e.g.}, AI agents~\cite{guo2024large}). As the scaling law gradually slows down, the test-time scaling~\cite{muennighoff2025s1} in LLM reasoning is emerging as a powerful tool in enhancing the model capabilities, especially in solving complex problems(\textit{e.g.}, mission planning~\cite{guo2024large, ferrag2025llm, zhang2025web} and mathematical derivation~\cite{ahn2024large}), through step-by-step chain-of-thought generation~\cite{hao2024llm}. Moreover, some latest works point out that as the length of the chain-of-thought reasoning increases, the LLM capabilities, especially the mission planning(\textit{e.g.}, $8K$ reasoning length) and information search($5M$ search length) in AI agents~\cite{zhang2025web}, can be significantly improved.

To effectively support long-context reasoning, LLM providers have enhanced the long-context processing capabilities of their LLMs during pretraining, with context windows of over 100K tokens becoming a common standard (\textit{e.g.}, Kimi K2 with 128K~\cite{team2025kimi} and OpenAI o3 with 200K~\cite{gpt-o3}). 
Despite these algorithm advances, the computational and memory burden associated with Key-Value(KV) cache still prevents the efficient practical deployment.
KV cache is a fundamental component in the LLM, which effectively reduces the computation by reusing past key-value pairs, but introduces significant memory overhead, which is proportional to the context length.
Furthermore, during the autoregressive decoding, the generation of each new token requires reading the entire KV cache to compute attention weights, leading to severe latency overhead. 
For example, in the case of Llama3.1-8B~\cite{grattafiori2024llama} on a NVIDIA RTX 4090 GPU, generating a single token with a 16K context takes twice as long as generating a token with a 1k context, 
and theoretically generates 2GB of memory footprint for KV cache, which means that for an NVIDIA RTX 4090 GPU, only 3 requests can be processed in parallel at most shown in Figure~\ref{fig:pareto}. For the LLM cloud service vendors, the longer response and limited throughput will translate to higher infrastructure costs (\textit{e.g.}, energy and hardware consumption) and suboptimal user experiences~\cite{hong2024flashdecoding++}.

Consequently, many previous works have explored various techniques for KV cache optimization, particularly in resource-constrained environments by reducing the KV cache involved during inference, encompassing algorithm optimization (\textit{e.g.}, permanent eviction~\cite{beltagy2020longformer, streamingllm} and dynamic selection~\cite{tang2024quest, liu2024clusterkv, sun2024shadowkv} of KV cache), system enhancement (\textit{e.g.}, customized CUDA kernel design~\cite{tang2024quest, liu2024clusterkv, sun2024shadowkv}). These algorithms establish a paradigm centered on layer-wise retrieval operation during the \textit{decoding} phase shown in Figure~\ref{fig:overview}(a). LLM inference can be divided into two phases, the \textit{prefill} and \textit{decoding} phase detailed in Section~\ref{sec:back}. Most previous works~\cite{liu2024clusterkv, sun2024shadowkv, tang2024quest} preprocess KV cache upon completion of the \textit{prefill} phase, and the corresponding retrieval algorithms retrieve a subset of preprocessed KV cache for each generation during \textit{decoding} phase. 

However, the core trade-off of this paradigm is its departure from mathematical equivalence. By selectively computing attention over a fraction of the context, these methods inherently introduce computational shortcuts that can lead to a degradation in model accuracy. Therefore, as illustrated in Figure~\ref{fig:pareto}(a)(b), two Pareto frontiers in long-context input and reasoning scenarios are established, forcing a compromise between inference speed and model accuracy. Despite this paradigm performing well in the long-context input scenario, it still suffers from the following critical limitations during the \textit{decoding} phase if directly applied in the long-context reasoning scenario.

\textbf{\textit{Challenge-1}: Time-consuming layer-wise retrieval operation.}
Figure~\ref{fig:overview}(a) shows that the algorithm paradigm needs to perform the retrieval over the KV cache and load the corresponding KV pairs based on the retrieval result before attention computation in each layer, resulting in the sequential dataflow due to data dependency. This serialization introduces substantial synchronization overhead, breaking the natural overlap between computation and memory access in the original pipeline. Furthermore, the retrieval operation is repeated in each layer during decoding, and thus the overhead scales linearly with model depth and quickly becomes bottleneck(up to $60\%$ latency) shown in Figure~\ref{fig:overview}(a).

\textbf{\textit{Challenge-2}: Complete retention of the newly generated KV cache.}
Existing works designed for the long-context input scenario preprocess the KV cache by complex and time-consuming algorithms(\textit{e.g.}, clustering~\cite{liu2024clusterkv} and quantization~\cite{sun2024shadowkv}) during the \textit{prefill} phase(\textit{i.e.}, the KV cache of the prompt), and only retrieve the preprocessed KV cache and completely retain the newly generated KV pair during the \textit{decoding} phase to avoid the repeated preprocessing shown in Figure~\ref{fig:overview}(a). With the substantial retrieval overhead in each layer, performance thus degrades greatly in the long-context reasoning scenario, even worse than full attention(\textit{i.e.}, FlashInfer~\cite{yeflashinfer}), as shown in Figure~\ref{fig:pareto}(b).

\begin{figure}[!t]
    \centering
    \includegraphics[width=0.48\textwidth]{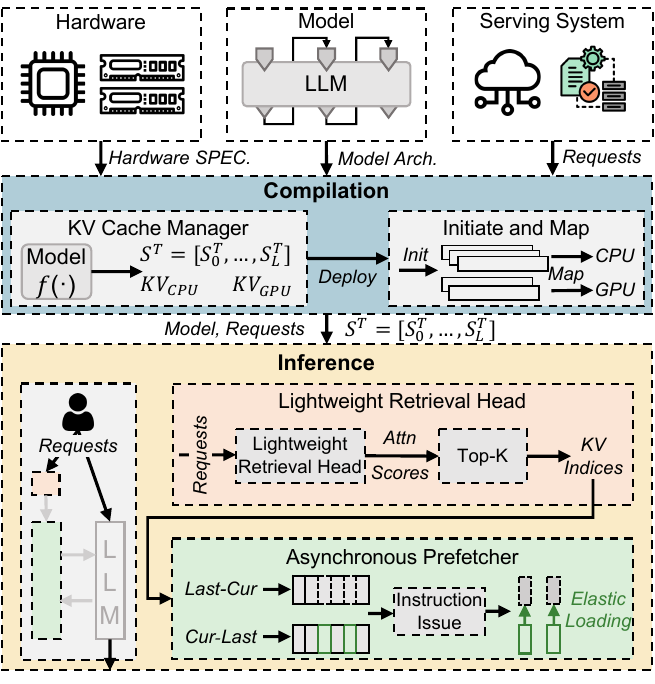}
    \vspace{-20pt}
    \caption{Architecture of \we.}
    \vspace{-10pt}
    \label{fig:arch}
\end{figure}

\textbf{\textit{Challenge-3}: Performance degradation with a tiny increase in sequence length.} In resource-constrained environments(\textit{e.g.}, low-end GPU with limited memory in edge and high-end GPU with multi-requests in cloud), KV cache tends to be offloaded to the lower-tier memory (\textit{e.g.}, from GPU HBM to CPU DRAM). However, existing systems determine the offloading strategy that either fully offloading or never offloading (\textit{e.g.}, ClusterKV~\cite{liu2024clusterkv}) before inference. Due to the inference dynamics in LLM reasoning, the predetermined strategy cannot adapt to the increasing sequence length during the autoregressive decoding, resulting in $>80\%$ performance degradation with a tiny increase in sequence length.

In this paper, we point out that the core objective of the retrieval algorithms is to align with the LLM, especially in the information focus, and the retrieval accuracy directly influences the LLM performance($>10\%$ accuracy gap between Quest~\cite{tang2024quest} and ClusterKV~\cite{liu2024clusterkv} using two different algorithms). 
Inspired by the objective of alignment in the output distribution in the LLM knowledge distillation~\cite{xu2024survey}, we consider that due to the homology between the distilled LM and the original LLM, the information they focus on (\textit{i.e.}, the important tokens) exhibits a high degree of similarity given the same inputs, and we also analyze this similarity through the mutual information~\cite{kraskov2004estimating} and the data processing inequality~\cite{beaudry2011intuitive} in information theory~\cite{shannon1948mathematical}.
Therefore, we propose \textbf{a novel paradigm that leverages a DLM as the retrieval algorithm to efficiently retrieve important information focus} shown in Figure~\ref{fig:overview}(b). Based on the insight, we present \we, an algorithm and system co-design for speculative context sparsity in long-context reasoning. The contributions of \we can be summarized into three levels as follows.

\textbf{(1) Lightweight retrieval head design at the algorithm level.}
Based on the insight mentioned above, we integrate a DLM before the LLM inference shown in Figure~\ref{fig:overview}(c)-C1, and explore the similarity of the focused tokens between the DLM and the original LLM based on the attention weights from two mapping dimensions, head-level and batch-level. Statistical data shows that there exists a higher similarity in the head-level dimension. Therefore, we design a lightweight retrieval head based on the head-level attention weights by pruning the redundant operations in DLM, achieving $>90\%$ parameter reduction.

\textbf{(2) Asynchronous prefetch dataflow via elastic loading at the system level.}
We further point out that, different from the existing works, \we selects the important KV pairs before the LLM inference through the lightweight retrieval head, eliminating the data dependency between the KV retrieval and loading during inference. Therefore, we design an asynchronous KV cache prefetch dataflow shown in Figure~\ref{fig:overview}(c)-C2. The dataflow only requires several lines of code about KV positions modification on the original LLM pipeline. Furthermore, we observe that the retrieval results between adjacent token generation are similar, and thus propose an elastic loading strategy into the dataflow, which only loads the different KV pair required by the current generation, successfully reducing data transfer by up to $90\%$.

\textbf{(3) Adaptive memory management at the compilation level.}
The critical path of LLM inference in resource-constrained environments is dominated by the latency of CPU-GPU data transfer. We develop a theoretical memory overhead model that considers LLM, hardware, and workload to optimize memory usage and inference latency by maximizing the GPU memory utilization. Guided by the model, we propose an adaptive memory management system shown in Figure~\ref{fig:overview}(c)-C3, which adaptively allocates memory to maximize the inference speed with increasing sequence length in LLM reasoning. 

The architecture of \we is shown in Figure~\ref{fig:arch}. \we begins when receiving the inference workload(\textit{e.g.}, requests) processed by the serving system. In the compilation stage, the adaptive memory management system calculates the sequence length thresholds based on the theoretical model and initializes the memory for the KV cache. During autoregressive inference, the lightweight retrieval head aims to identify critical KV pairs in all KV cache and obtain their indices. These indices are immediately fed to the asynchronous prefetcher for difference calculation, kicking off KV prefetching with elastic loading in parallel with the original LLM inference to enable the overlap of GPU computation and CPU-GPU data transfer.

\begin{figure*}[!t]
    \centering
    \includegraphics[width=0.98\textwidth]{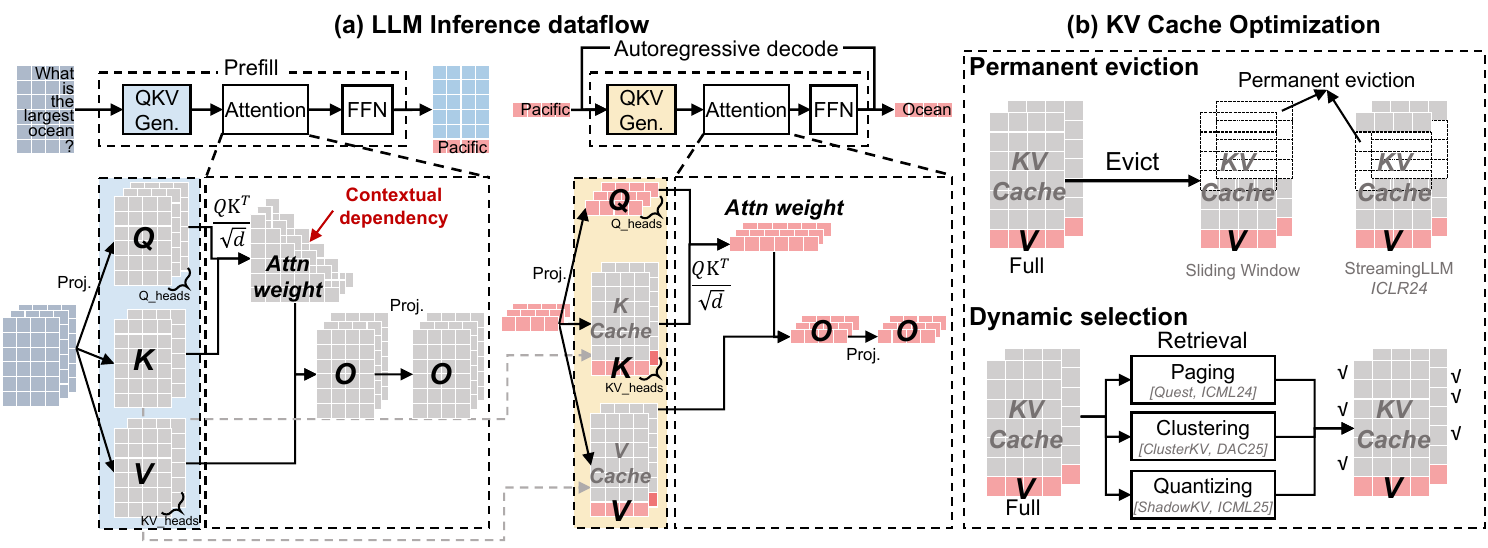}
    \vspace{-10pt}
    \caption{(a) Inference dataflow of LLM. (b) Existing works on KV cache optimization.}
    \vspace{-10pt}
    \label{fig:background}
\end{figure*}

We deploy and evaluate \we in two resource-limited environments, a low-end GPU with limited memory in edge and a high-end GPU with multiple requests in cloud, targeting long-context input and reasoning scenarios. Extensive experiments demonstrate that \we achieves $24.89\times$ throughput improvement in the cloud environment and $10.06\times$ speedup in the edge environment with negligible accuracy loss, pushing the Pareto frontier of accuracy and throughput for long-context input and reasoning scenarios.


\section{Background and Related Work} \label{sec:back}

\subsection{Large Language Model} \label{sec:back:LLM}
Figure~\ref{fig:background}(a) shows that LLM inference is composed of two phases, \textit{prefill} and \textit{decoding} phase. The \textit{prefill} phase processes the prompt to generate the first token and caches its key-value pairs. Subsequently, the \textit{decoding} phase uses the KV cache to generate the new token autoregressively and appends the new key-value pair to the KV cache. Nowadays, mainstream LLMs select the Transformer decoder~\cite{vaswani2017attention} as the backbone layer, which primarily includes two modules, the attention mechanism and the feed-forward network(FFN). The attention mechanism requires that the current token generation is solely dependent on previous tokens. The FFN aims to capture deeper features and handle nonlinear relationships.

\subsection{KV Cache Optimization} \label{sec:back:KV}
As illustrated in Figure~\ref{fig:background}(a), to reduce computation, existing LLM inference systems leverage the KV cache to store the keys and values generated from this previous content, but introduce the memory overhead that scales linearly with the context length (\textit{e.g.}, 4GB memory footprint with 32K context in Llama3.1-8B~\cite{grattafiori2024llama}), posing significant challenges in resource-constrained environments. 

Owing to the \textit{softmax} operation in attention described in Equation~\ref{eq:attn}, the attention weights exhibit approximate sparsity (\textit{i.e.}, many values are close to zero). Capitalizing on this phenomenon, many techniques emerged to optimize the KV cache, such as permanent eviction and dynamic selection. 
\begin{equation}
    Attn\_weight = softmax(\frac{QK^T}{\sqrt{d}})
\label{eq:attn}
\end{equation}


\textbf{Permanent eviction.} 
Sliding Window~\cite{beltagy2020longformer} is a typical representative of permanent eviction and is still used in some industrial LLM deployments(\textit{e.g.}, Gemma 3~\cite{team2025gemma}). It retains only a fixed number of the most recent KV pairs(\textit{i.e.}, ``window") and evicts the farthest ones as new tokens are generated(\textit{i.e.} ``sliding"). While this approach ensures a constant memory for the KV cache, it discards too much historical context, resulting in significant accuracy loss. StreamingLLM~\cite{streamingllm} represents a notable optimization on this paradigm. It builds on the insight that, due to the nature of the \textit{softmax}, the initial few tokens accumulate a wealth of information, called ``attention sink". Therefore, in addition to the sliding window, StreamingLLM perpetually retains these crucial initial KV pairs to improve the model accuracy.

\textbf{Dynamic selection.}
To address the significant accuracy degradation caused by irreversible information loss in permanent eviction, some works~\cite{tang2024quest, liu2024clusterkv, sun2024shadowkv} propose the dynamic selection, which retains the entire KV pairs or offloads them to lower-tier memory(\textit{e.g.}, CPU DRAM) in resource-constrained environments, and retrieves the necessary KV pairs based on the input during inference. In order to minimize the retrieval overhead, most works require preprocessing the KV cache (\textit{e.g.}, paging~\cite{tang2024quest}, clustering~\cite{liu2024clusterkv}, and quantization~\cite{sun2024shadowkv}). Given the substantial overhead of the preprocessing, most works only preprocess the KV cache of the input prompt after the \textit{prefill} phase, and only retrieve the preprocessed KV cache during the \textit{decoding} phase with the retention of newly generated KV pairs. 
Quest~\cite{tang2024quest} is a representative work in dynamic selection, which partitions the KV cache into pages and creates a page vector by taking the element-wise maximum and minimum values. During retrieval, importance scores are computed only for these page vectors to select the Top-K pages. Subsequently, all KV pairs within the selected pages are loaded for computation. ClusterKV~\cite{liu2024clusterkv} improves upon Quest by employing clustering to categorize the KV cache. It uses the cluster centroids as the cluster vectors for the importance calculation, leading to a notable accuracy improvement. Similarly, the ShadowKV~\cite{sun2024shadowkv} quantizes the key cache and computes attention between the query and the quantized keys. Based on the results, it selects the important KV pairs for computation. A common characteristic of all these approaches is their reliance on preprocessing the KV cache. This requirement is ill-suited for the long-context reasoning scenario, where the KV cache continuously grows during the \textit{decoding} phase, making repeated preprocessing computationally expensive. In this paper, \we aims to achieve the efficient long-context reasoning of LLM through the lightweight retrieval head on raw KV cache without complex preprocessing.

\subsection{Knowledge Distillation in LLMs} \label{sec:back:KD}
Knowledge Distillation is a typical technique to address the challenge of deploying the LLMs in some resource-constrained scenarios. Its primary goal is to compress a large ``teacher" LLM  into a smaller, more efficient ``student" LLM while preserving high performance. The student LLM learns to mimic the outputs of the teacher LLM to achieve the alignment of the probability distributions by minizing the Kullback-Leibler Divergence~\cite{kullback1951information} formulated as follows.
\begin{equation}
    D_{KL}(P_T||P_S) = \sum_{i} P_T(x_i)log(\frac{P_T(x_i)}{P_S(x_i)})
    \label{eq:KL}
\end{equation}
The $P_T$ denotes the probability distribution of the teacher LLM, and the $P_S$ denotes the probability distribution of the student model.
Recently, knowledge distillation is further used to accelerate LLM inference through speculative decoding. The EAGLE family~\cite{li2024eagle-1,li2024eagle-2,li2025eagle-3} is a representative work. It leverages a distilled small language model to autoregressively generate draft tokens, which are then fed into the LLM for parallel verification. Since the training objective of the distilled model is to align its output distribution with that of the LLM, the number of tokens passing verification is often greater than one, allowing the LLM to generate multiple tokens in a single forward inference.

\section{Motivation} \label{sec:insight}

\subsection{Two Core Questions of KV Selection} \label{sec:insight:questions}

\textit{\textbf{Question: What is the essential objective of retrieval algorithms? }}

\textit{\textbf{Answer:} The retrieval algorithms aim to efficiently align with the intrinsic properties on the LLM contextual focus.}

\textit{\textbf{Analysis:}} As mentioned in Section~\ref{sec:back:KV}, permanent eviction strategies~\cite{beltagy2020longformer,streamingllm} are typically informed by coarse-grained statistical and theoretical analysis of the attention mechanism. These works reveal intrinsic, input-agnostic properties of LLMs, such as the consistent focus on local context or specific absolute positions (\textit{e.g.}, the initial tokens~\cite{streamingllm}), leading to the design of fixed retrieval algorithms that are independent of the input query. In contrast, dynamic selection strategies~\cite{tang2024quest, liu2024clusterkv, sun2024shadowkv} are based on fine-grained experimental analysis that the focus is highly dynamic and content-dependent. By leveraging the intrinsic properties of LLMs(\textit{e.g.}, representational similarity~\cite{liu2024clusterkv} and low-rank characteristics~\cite{sun2024shadowkv}), these works propose query-aware retrieval algorithms. As illustrated in Figure~\ref{fig:overview}(b), we point out that the core of the retrieval algorithms is to first identify the intrinsic properties of the LLM on the contextual focus, and then align with these properties in an efficient way. The alignment degree between the retrieval algorithm and LLM decides retrieval accuracy, proportional to the model accuracy.

\textbf{\textit{Question: Why do the existing retrieval algorithms need complex and time-consuming preprocessing?}}

\textit{\textbf{Answer:} The primary purpose of preprocessing is to mitigate the computational overhead during retrieval.}

\textit{\textbf{Analysis:}} To retrieve the important KV pairs, most existing retrieval algorithms in each layer take matrix multiplication of $Query\in R^{bsz \times heads\times dim}$ with $Keys_{candidate} \in R^{dim \times heads \times len_{keys}}$ to get the importance scores, and then select the Top-K candidates for the final computation. Therefore, the retrieval overhead($O_{tot}$) in a single LLM inference can be defined as follows.

\begin{equation}
    O_{tot} = layers  \times bsz \times  heads\times dim \times len_{keys} \times O_{mul}
\label{eq:overhead}
\end{equation}
As mentioned in Section~\ref{sec:back:KV}, Quest and ClusterKV leverage preprocessing algorithms(\textit{e.g.}, paging and clustering) which select a single vector as the representative of several Keys, to reduce the length of candidate keys($len_{keys}$). ShadowKV reduce the single multiplication overhead($O_{mul}$) by quantizing the key vectors to low bit level. From Equation~\ref{eq:overhead}, if without preprocessing, the retrieval overhead is equivalent to the attention weights computation in Equation~\ref{eq:attn}, losing the meaning of KV selection.


\subsection{Key Insight and Theoretical Analysis} \label{sec:insight:analysis}
\textbf{Key Insight.}
Inspired by the alignment objective of knowledge distillation in LLM and its wide application across various domains(\textit{e.g}, speculative decoding~\cite{li2024eagle-1, li2024eagle-2, li2025eagle-3} and early exiting~\cite{xu2025specee}), we consider that the goal of DLM in these works is to generate the probability distribution that resembles the original LLM. From the perspective of information, we intuitively consider that if the probability distribution is nearly the same, the contextual information focus(\textit{i.e.}, the tokens contribute most to the result) in the DLM and the original LLM must be highly similar. Otherwise, any significant information discrepancy would prevent the alignment in the probability distribution. 

\textbf{Theoretical Analysis.}
As mentioned in Section~\ref{sec:back:KD}, the objective of knowledge distillation in LLMs is to minimize the KL divergence of the probability distributions in Equation~\ref{eq:KL}.  We consider that this inherently requires the student model to learn the context information extraction strategy similar to that of the teacher model. This insight can be analyzed through mutual information~\cite{kraskov2004estimating} and the data processing inequality~\cite{beaudry2011intuitive} in information theory~\cite{shannon1948mathematical}. Mutual information($I(X;Y)$) measures the dependence between two variables. For a well-trained teacher model($T$), there exists high mutual information between the output probability distribution ($P_T$) and the input context($C$)(\textit{i.e.}, $I(C;P_T)$ is large). This means that the teacher's output is highly dependent on the context and not random guessing. Moreover, the information flow from the context($C$) through the internal representation($R_S$) of the student model($S$) to output probability distribution ($P_S$ forms a Markov chain~\cite{markov1906rasprostranenie}:
\begin{equation}
    C\rightarrow R_S \rightarrow P_S
\end{equation}
According to the DPI, we can get 
\begin{equation}
    I(C,P_S) \leq I(C,R_S)
\end{equation}
This indicates that the amount of information about the context contained in the output cannot exceed the information captured by its internal representation. The distillation process drives $P_S \rightarrow P_T$ by minimizing $D_{KL}(P_T||P_S)$. Since $P_T$ has high mutual information with $C$, a successful distillation will ensure that $P_S$ also exhibits high mutual information with $C$, i.e., $I(C,P_S)\rightarrow I(C;P_T)$. To achieve a high level of $I(C,P_S)$, the DPI dictates that the student model must learn to generate an internal representation($R_S$) that also captures significant contextual information in $C$, ensuring that $I(C,R_S) \geq I(C,P_S)$. Therefore, the student model must extract the contextual information that the teacher model deems important.

Building on the insight and analysis above, we propose \textbf{a novel paradigm that leverages the DLM of the original LLM as the retrieval algorithm}. This paradigm can transfer the information focus from the DLM to the original LLM during inference, eliminating layer-wise time-consuming retrieval detailed in Section~\ref{sec:intro} and complex preprocessing mentioned above, and thus effectively supporting the long-context reasoning scenario.

\begin{figure*}[!t]
    \centering
    \includegraphics[width=0.98\textwidth]{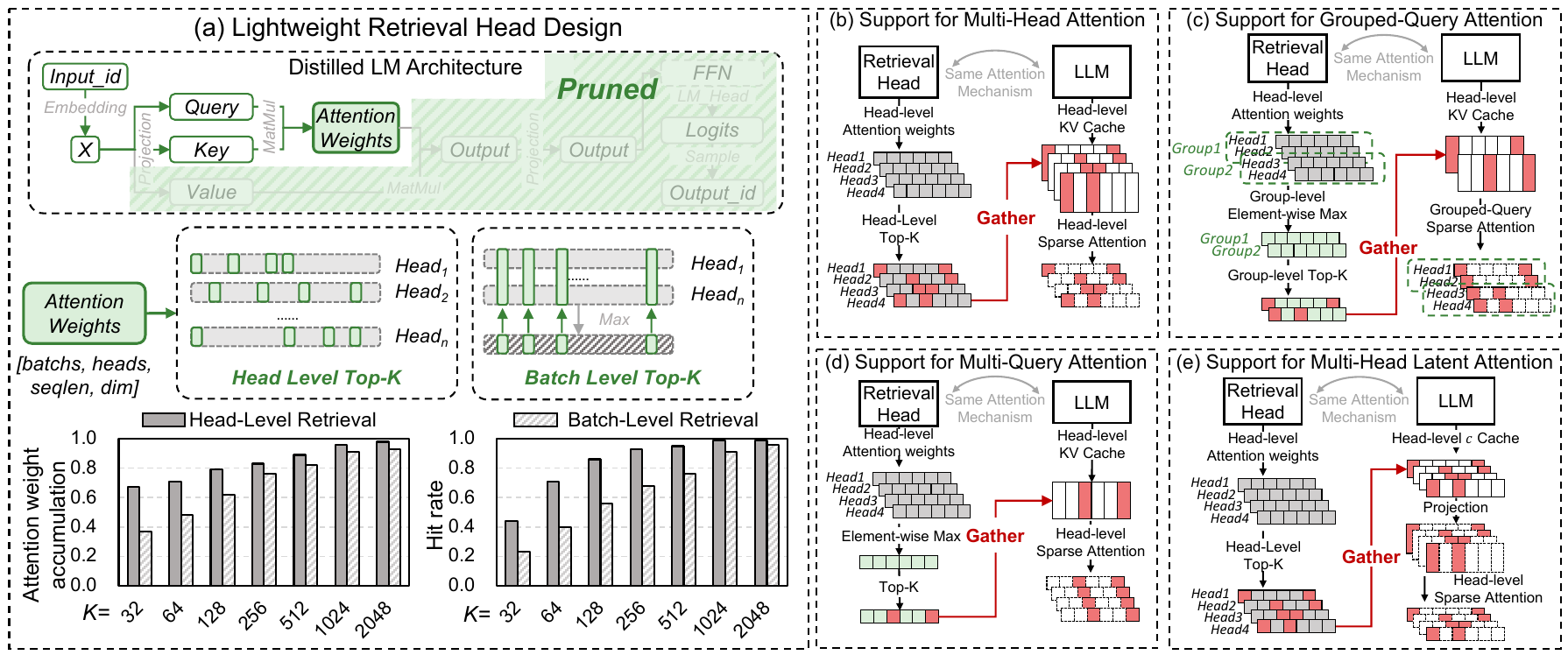}
    \vspace{-10pt}
    \caption{(a) We design the lightweight retrieval head by pruning redundancy and adopt the head-level attention weights for selection. (b)$\sim$(e) The detailed implementations of four attention mechanisms supported by the lightweight retrieval head.}
    \vspace{-10pt}
    \label{fig:tech1}
\end{figure*}


\section{Lightweight Retrieval Head Design} \label{sec:T1}

\subsection{Challenge: Time-consuming DLM}
Based on the key insight mentioned above, we deploy the DLM before the original LLM to capture globally important tokens shown in Figure~\ref{fig:arch}. In this paper, we utilize the DLM provided by EAGLE-3~\cite{li2025eagle-3}, which has the complete LM architecture(including \textit{tokenizer}, \textit{embedding} and \textit{LM\_Head}) with a single Transformer decoder layer. As shown in Figure~\ref{fig:arch}, the DLM processes the same inputs as the LLM, and performs complete inference with the full KV cache, resulting in $\sim20\%$ additional overhead, especially for the LLM with large vocabulary(\textit{e.g.}, $>1.2\times10^5$ tokens in Llama3-8B). Therefore, the key issue is \textbf{how to design a lightweight retrieval algorithm based on the DLM} to minimize the overhead.

\subsection{Insight and Analysis: Redundant Operation} \label{sec:T1:insight}
We further point out that the role of DLM is primarily to retrieve important tokens, which are often determined by the attention weights. Therefore, we conducted experiments on the similarity between the DLM and the original LLM from two mapping dimensions in attention weights, batch-level and head-level, shown in Figure~\ref{fig:tech1}(a). The batch-level retrieval adopts a coarse-grained approach, retaining a single set of important tokens that apply to all attention heads. In contrast, the head-level retrieval is more fine-grained, retaining different important tokens for each attention head. As illustrated in Figure~\ref{fig:tech1}, the head-level retrieval exhibits a higher similarity of the important tokens and higher hit rate of the generated tokens. Therefore,  we only need operations related to the calculation of attention weights (\textit{e.g.}, Query and Key generation), while other operations are redundant.

\subsection{Approach: Lightweight Retrieval Head}
Building on the insight and analysis, we design the lightweight retrieval head based on the DLM. The retrieval head supports three mainstream LLM attention mechanisms (\textit{i.e.}, Multi-Head Attention(MHA), Grouped-Query Attention(GQA), Multi-Query Attention(MQA), and Multi-Head Latent Attention(MLA)). The implementation details are as follows.

\textbf{Implementation Details.}
As illustrated in Figure~\ref{fig:arch}, we deploy the retrieval head before the original LLM and process the same input as the LLM.  This retrieval head retains the essential components of DLM provided by EAGLE-3~\cite{li2025eagle-3}, the embedding module and the QK projection weights. Although the original DLM only supports 2k context length, we enable it to process long context using the training-free method provided by YaRN~\cite{pengyarn}. During the inference, the retrieval head maintains a full Key (K) cache and calculates attention weights after the QK projection. Based on the analysis in Section~\ref{sec:T1:insight}, we perform the head-level retrieval of important tokens based on the attention weights and feef the selected tokens into the original LLM inference. The implementation of the head-level retrieval tailored for the different attention mechanisms is as follows.

\textbf{Support for MHA.}
MHA was once a mainstream attention mechanism adopted by many LLMs(\textit{e.g.}, Llama-2~\cite{touvron2023llama}). The number of heads for Keys($K$) and Values($V$) is the same as that of Queries($Q$) in Figure~\ref{fig:background}(a). Since the attention mechanism of the retrieval head is the same as that of the original LLM, the retrieval head selects important tokens at the head level based on the attention weights shown in Figure~\ref{fig:tech1}(b). The selected tokens are then mapped to the attention computation of the original LLM by using the \texttt{torch.gather} operation to load the important KV cache into different heads.

\textbf{Support for GQA.}
GQA is introduced to optimize the substantial KV cache overhead of MHA, and most mainstream LLMs(\textit{e.g.}, Llama3~\cite{grattafiori2024llama} and Qwen3~\cite{yang2025qwen3}) have updated to GQA. As illustrated in Figure~\ref{fig:tech1}(c), GQA divides the query heads into groups, where all heads in the group share the same KV cache. Consequently, the number of heads in the KV cache is reduced to $\frac{1}{\alpha}$ of the query heads, where $\alpha$ is the number of groups. For computational convenience, the KV heads are often repeated $\alpha$ times before the attention calculation, resulting in attention weights with the same number of heads as the query. This thus creates the mismatch between the attention weights of the retrieval head and the physical KV cache of original LLM in head numbers. To address this, as shown in Figure~\ref{fig:tech1}(c), we apply an element-wise maximum operation along the hidden dimension within the same group of heads in the attention weights, to generate the group-level attention weights. We then take the group-level attention weights for important token selection and subsequent operations, which are similar to MHA.

\textbf{Support for MQA.}
MQA divides all heads of the query into a single group, where all heads share the same KV cache. Therefore, the implementation of \we in MQA is similar to that in GQA shown in Figure~\ref{fig:tech1}(d), \textit{i.e.}, $n$ is changed to the number of all heads. 

\textbf{Support for MLA.}
MLA is a novel variant of MHA employed in a new series of models(\textit{e.g.}, DeepSeek-V3/R1~\cite{liu2024deepseek} and Kimi-K2~\cite{team2025kimi}). Instead of caching the full Key-Value pairs, MLA caches a lower-dimensional latent representation, denoted as $c$. During computation, the $c$ is mapped to a higher dimensional space for the attention calculations. Since MLA does not reduce the number of attention heads, our retrieval remains similar to that in MHA. The primary difference lies that only the selected $c$ cache is subjected to the increase in dimension as shown in Figure~\ref{fig:tech1}(e).

\section{Asynchronous Prefetch Dataflow}

\subsection{Motivation: Data independence}
\begin{figure}[!t]
    \centering
    \includegraphics[width=0.48\textwidth]{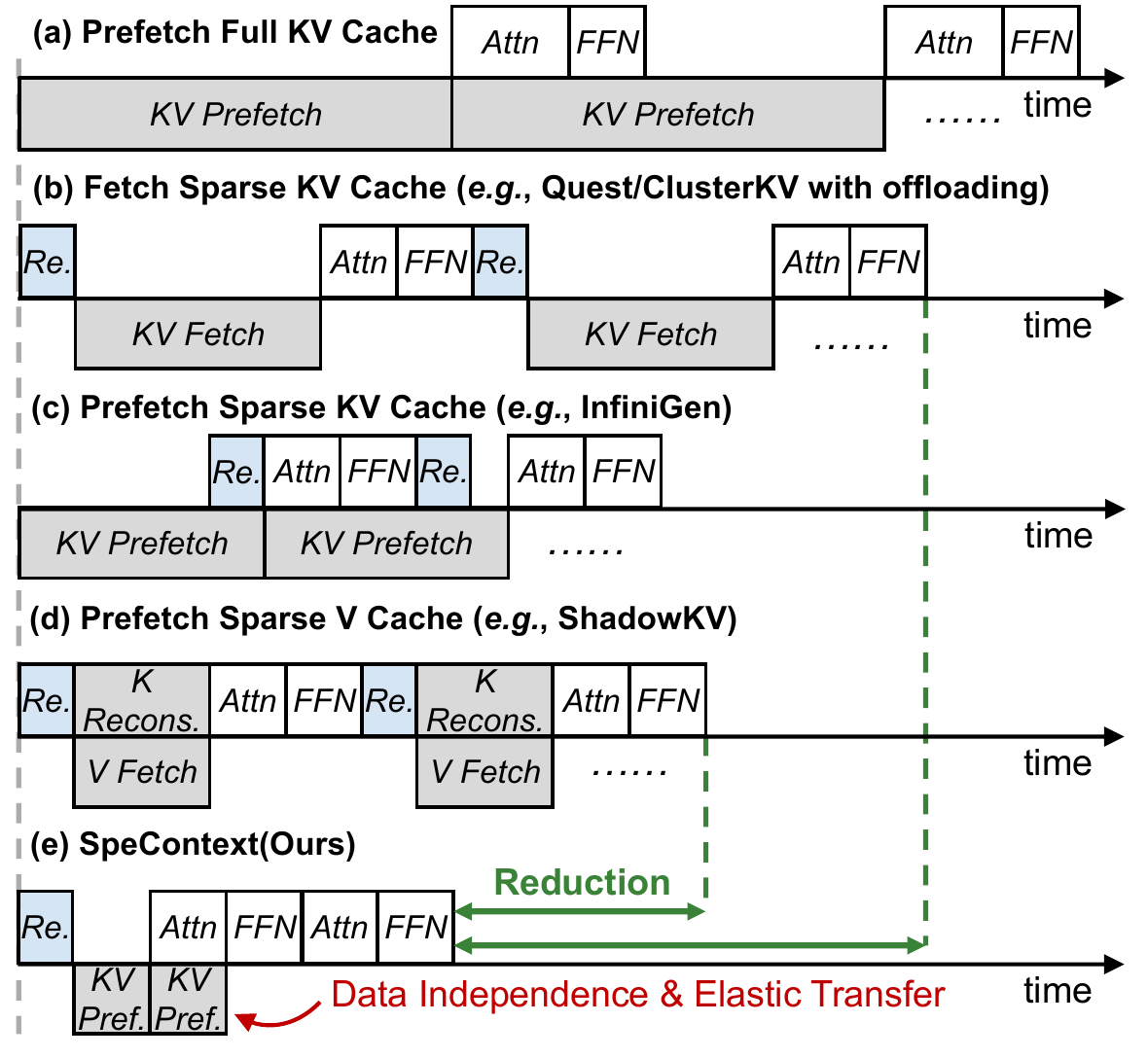}
    \vspace{-10pt}
    \caption{Elastic loading effectively reduces the KV transfer, making \we outperform previous works. }
    \vspace{-16pt}
    \label{fig:tech2}
\end{figure}

As previously mentioned in Section~\ref{sec:intro},
inference engines will offload the KV cache to lower-tier memory in resource-constrained environments.
As illustrated in Figure~\ref{fig:tech2}, existing KV cache retrieval works must load the required KV cache based on retrieval results for attention computation in each layer. This design introduces the synchronization and control caused by data dependencies. As mentioned in Section~\ref{sec:insight:analysis}, the lightweight retrieval head in is deployed before LLM inference and dependent solely on the LLM input. 
eliminating the data dependency mentioned above. Consequently, we further propose the asynchronous dataflow through multiple CUDA streams, enabling concurrent execution of computation and KV cache prefetching shown in Figure~\ref{fig:overview}(c)-C2.

\begin{figure}[!t]
    \centering
    \includegraphics[width=0.48\textwidth]{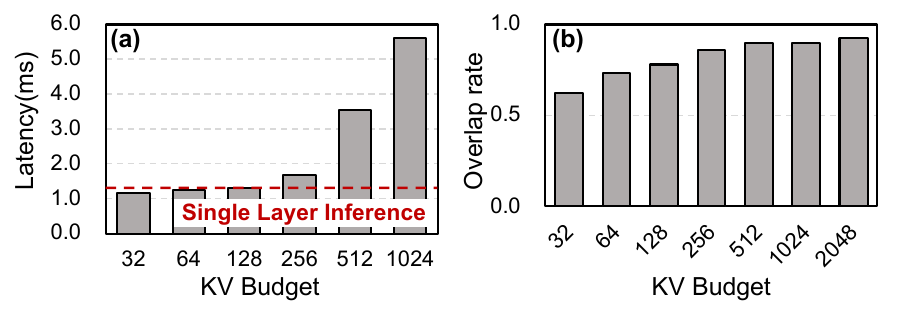}
    \vspace{-15pt}
    \caption{(a) The latency of prefetching with different KV budget and a LLM layer inference. (b) Overlap rate of selected tokens in adjacent generation with different KV budget.}
    \vspace{-15pt}
    \label{fig:overlap}
\end{figure}

\subsection{Challenge: Heavy Data transfer}
However, due to the combination of limited memory bandwidth and the immense computational power of GPUs, a significant imbalance arises. 
As illustrated in Figure~\ref{fig:overlap}(a), for the large KV budget(\textit{i.e.}, self-determined amount of KV cache for loading), the data transfer latency far exceeds the LLM inference latency
As a result, in resource-constrained scenarios, the end-to-end inference latency becomes dominated by the I/O for loading the KV cache. 
Therefore, the key challenge is \textbf{how to reduce the data transfer time (\textit{i.e.}, minimize the volume of KV cache loaded) without sacrificing accuracy}.

\subsection{Insight: Contextual Similarity}
Inspired by the contextual similarity explored in early exiting~\cite{xu2025specee} and sparse activation~\cite{mirzadehrelu}, we conduct experiments to explore the relationship of the selected tokens between two adjacent token generation. As illustrated in Figure~\ref{fig:overlap}(b), statistical analysis reveals the high overlap($>80\%$) in the important token selection between adjacent generation. This implies that for the subsequent generation, only about $20\%$ of the KV cache on GPU requires to be updated. Consequently, we can maintain the accuracy by loading only $20\%$ updating KV cache, effectively reducing the data transfer volume.

\subsection{Approach: Elastic Loading}
Based on the contextual similarity, we propose the elastic loading strategy and integrate it into the asynchronous dataflow. Its objective is to reuse the KV cache already resident on the GPU from the previous generation and fetch only those not yet present. This strategy can be implemented with minimal code modifications to the existing asynchronous dataflow. The implementation details are as follows. We denote the set of important token indices in last generation as $S_{last}$. And we obtain the indices $S_{now}$ for current generation by the retrieval head. The set of KV cache indices to be updated on GPUs can be calculated by the set difference $S_{last} - S_{now}$ while the KV cache indices for elastic loading are calculated by $S_{now} - S_{pre}$. Because we maintain a fixed KV budget(\textit{i.e.}, $|S_{last}| = |S_{now}|$), it follows that $|S_{last} - S_{now}| = |S_{now} - S_{last}|$. Then $S_{last}$ needs to be updated by $S_{now}$. Practically, we perform in-place updates for the required KV loading through \texttt{Tensor.copy\_()}.

\section{Adaptive Memory Management} \label{sec:T3}

\subsection{Motivation: Performance Degradation}
As mentioned in Section~\ref{sec:intro}, most existing works designed for long-context input scenario determines the KV cache management strategy that whether to store the entire cache in GPU HBM or offload it to CPU DRAM before LLM inference. However, unlike the long-context input scenario, the sequence length exhibits is dynamic and unpredictable in the long-context reasoning scenario because the inference termination is completely determined by the LLM itself. Consequently, as shown in Figure~\ref{fig:overview}(a), even a tiny increase in task workload (\textit{e.g.}, the longer context length and more requests) can trigger a complete offload of the entire KV cache to the CPU, leading to $>80\%$ performance degradation.

\begin{table}[!t]
\small
\centering
\caption{Symbols mentioned in Section~\ref{sec:T3} and description.}
\vspace{-10pt}
\label{tab:notation}
\begin{tabular}{@{}ccc@{}}
\toprule
Category             & Symbol               & Description          \\ \midrule
\multirow{10}{*}{Model}                        &$M_O$                      &Memory size of original LLM \\
                     &$M_D$                      & Memory size of DLM    \\
                     &$L$                      & Number of layers in LLM  \\
                     &$D$                      & Head dimension in LLM \\
                     &$H$                      & Number of KV heads in LLM \\
                     &$S$                      & Current sequence length \\
                     &$B$                      & KV cache retrieval budget \\
                     &$L_{CPU}$                      & Number of layers of KV cache on CPU                    \\
                     &$L_{GPU}$                      & Number of layers of KV cache on GPU                      \\ 
                     &$\alpha$                      & Groups of attention heads                      \\ \midrule
Hardware  &$Mem_{GPU}$                      & Size of GPU global memory                     \\ \midrule
Workload                     &$R$                      &  Requests                    \\ \bottomrule
\end{tabular}
\vspace{-10pt}
\end{table}

\begin{algorithm}[!b]
\small
\caption{Sequence length threshold calculation in compilation}\label{alg:management_comp}
    \begin{algorithmic}[1]
      \REQUIRE The Symbols in Table~\ref{tab:notation}.
      \ENSURE The sequence length threshold list $S^T=[S^T_0,...S^T_L]$.
      \STATE $S_0^T \leftarrow \lfloor\frac{Mem_{GPU}-1.3\times(M_O+M_D)}{4 \times R\times H\times D\times(L+1+\alpha)}\rfloor$  \COMMENT{Place all KV cache on GPU}
      \FOR{$i \leftarrow 1$ to $L$}
      \STATE $S_i^T = \lfloor\frac{Mem_{GPU} - 1.3\times(M_O+M_D) - (i\times B)\times R\times H\times D}{4\times(L+1+\alpha-i) \times R\times H \times D}\rfloor$ \COMMENT{Place last i layers of KV cache on GPU}
      \ENDFOR
      \RETURN $S^T$
    \end{algorithmic}    
\end{algorithm}

\subsection{Approach: Adaptive Memory Management}

\textbf{Theoretical Model and Analysis.}
Inspired by a series of works on adaptive scheduling for resource allocation~\cite{xu2025specee, dai2025flashdecoding++next}, we develop a theoretical memory overhead model based on LLM architecture, hardware specifications and inference workload detailed in Table~\ref{tab:notation}, and further propose a novel adaptive memory management system. 


During LLM inference, additional memory, called runtime memory, is required to serve as a temporary buffer to store intermediate values (\textit{e.g.}, activations). We checked some references and found that runtime memory typically amounts to around 20\% to 30\% of the model size~\cite{dai2025flashdecoding++next, zhong2024distserve,VM-blog,VCF-blog}. This buffer is dynamically used and released during the LLM inference. As a result, we select $30\%$ of the model size as the runtime memory. As mentioned in Section~\ref{sec:T1}, the decode layer of DLM is consistent with the original LLM architecture and has only one layer, and due to the \texttt{repeat\_kv} operation in GQA or MQA, an additional buffer($S \alpha HD$) must allocated for computation. So the total number of layers in the KV cache is $(L+1+\alpha)$. Moreover, Due to the Key and Value with FP16 precision which is 2 byte per value, the coefficient of KV cache is $4$.
Therefore, we can calculate the total memory requirements for placing all KV cache the GPU as follows.
\begin{equation}
    M_{all}=M_{Model} +M_{KV}= 1.3(M_O+M_D)+4R(L+1+\alpha)SHD
\label{eq:TGPU}
\end{equation}

Based on the Equation~\ref{eq:TGPU}, if we keep all the data on the GPU, we need to make sure that $M_{all}$ < $Mem_{GPU}$.

For the resource-constrained environment(\textit{e.g.}, low-end GPU with limited memory(\textit{i.e.}, $Mem_{GPU}$ is insufficient) and high-end GPU with multi-requests(\textit{i.e.}, $M_{all}$ is too large)), it is necessary to split the KV cache across different memory tiers. Specifically, the KV cache of some layers($L_{GPU}$) should be stored on the GPU, while the KV cache of others($L_{CPU}$) are offloaded to the CPU. However, for the layers where the KV cache is offloaded to the CPU, it is still necessary to reserve a small GPU buffer to store KV cache budget($B$) loaded from the CPU for the computation. Therefore, the total memory requirements in this case can be calculated as follows:
\begin{equation}
    M_{part} = 1.3(M_O+M_D)+ 4R[(L_{GPU}+1+\alpha)S+(L_{CPU}B)]HD
\end{equation}

To maximize the utilization of the GPU memory, the theoretical optimization model thus can be abstracted as follows.
\begin{equation}
\begin{aligned}
    Max&(L_{GPU}); \\
    s.t. \ \ M_{part} &\leq Mem_{GPU}
\end{aligned}
\label{eq:model}
\end{equation}

\subsubsection{Implementation Details}

\begin{algorithm}[!t]
\small
\caption{Adaptive memory management in inference}\label{alg:management_infer}
    \begin{algorithmic}[1]
      \REQUIRE The sequence length threshold list: $S^T=[S^T_0,...S^T_L]$, Prompt: $input\_id$, LLM Layers: $Layer =[Layer_0,... Layer_{L-1}]$.
      \STATE $L_{CPU} \leftarrow 0$; $L_{GPU} \leftarrow L$ \COMMENT{Initiate $L_{CPU}$ and $L_{GPU}$}
      \STATE $S\leftarrow len(input\_id)$ \COMMENT{Initiate Sequence length $S$}
      \WHILE{True}
      \WHILE{$S\geq S^T_{L_{CPU}}$ and $L_{CPU} < L$}
      \STATE $KV\_Cache\_Offload(L-L_{CPU}-1)$ \COMMENT{Offload the KV cache of $Layer_{L-L_{CPU}-1}$}
      \STATE $L_{CPU} \leftarrow L_{CPU}+1$
      \ENDWHILE
      \STATE $input\_id = LLM(input\_id)$
      \STATE $S\leftarrow S+1$ \label{alg:line:9}
      \IF{$stop\_id$ in $input\_id$}
      \STATE Break
      \ENDIF
      \ENDWHILE
    \end{algorithmic}
\end{algorithm}

As mentioned above, the sequence length grows continuously during long-context reasoning. Following the objective of maximizing 
$L_{GPU}$ in Equation~\ref{eq:model}, we propose an adaptive memory management system that progressively offloads the KV cache of each LLM layer to the CPU as the context length increases during reasoning, thereby freeing additional GPU memory to store more KV cache in other layers. Our analysis indicates that during the inference, the primary factor influencing memory overhead is the sequence length. Capitalizing on this, we can pre-calculate the sequence length thresholds in compilation detailed in Algorithm~\ref{alg:management_comp}. The threshold $S_0^T$ represents that if we want to place all the KV cache on GPU, the current sequence length($S$) must be smaller than $S_0^T$. If $S>S_0^T$, we need to offload the KV cache of the final layer to CPU for more GPU memory.

During LLM inference, the adaptive memory management system will offload the KV cache of an additional layer to CPU DRAM at the exact time point based on these thresholds to maintain optimal memory usage. Algorithm~\ref{alg:management_infer} shows the details of LLM inference with adaptive memory management. For example, if the prompt length is between $S_2^T$ and $S_3^T$ at the beginning of LLM inference, the system will offload the KV cache of last two layers(\textit{e.g.}, the 31st and 32nd layer in Llama3-8B) to CPU DRAM and keep the KV cache of the left layers on GPU. As sequence length increases during inference(\textit{i.e.}, line~\ref{alg:line:9} in Algorithm~\ref{alg:management_infer}) and exceeds $S_3^T$, the management system will offload the KV cache of the 30th layer to CPU. With the adaptive memory management, we maximize the utilization of GPU HBM for better performance and convenient deployment.

\section{Evaluation}
\begin{figure}[!t]
    \centering
    \includegraphics[width=0.48\textwidth]{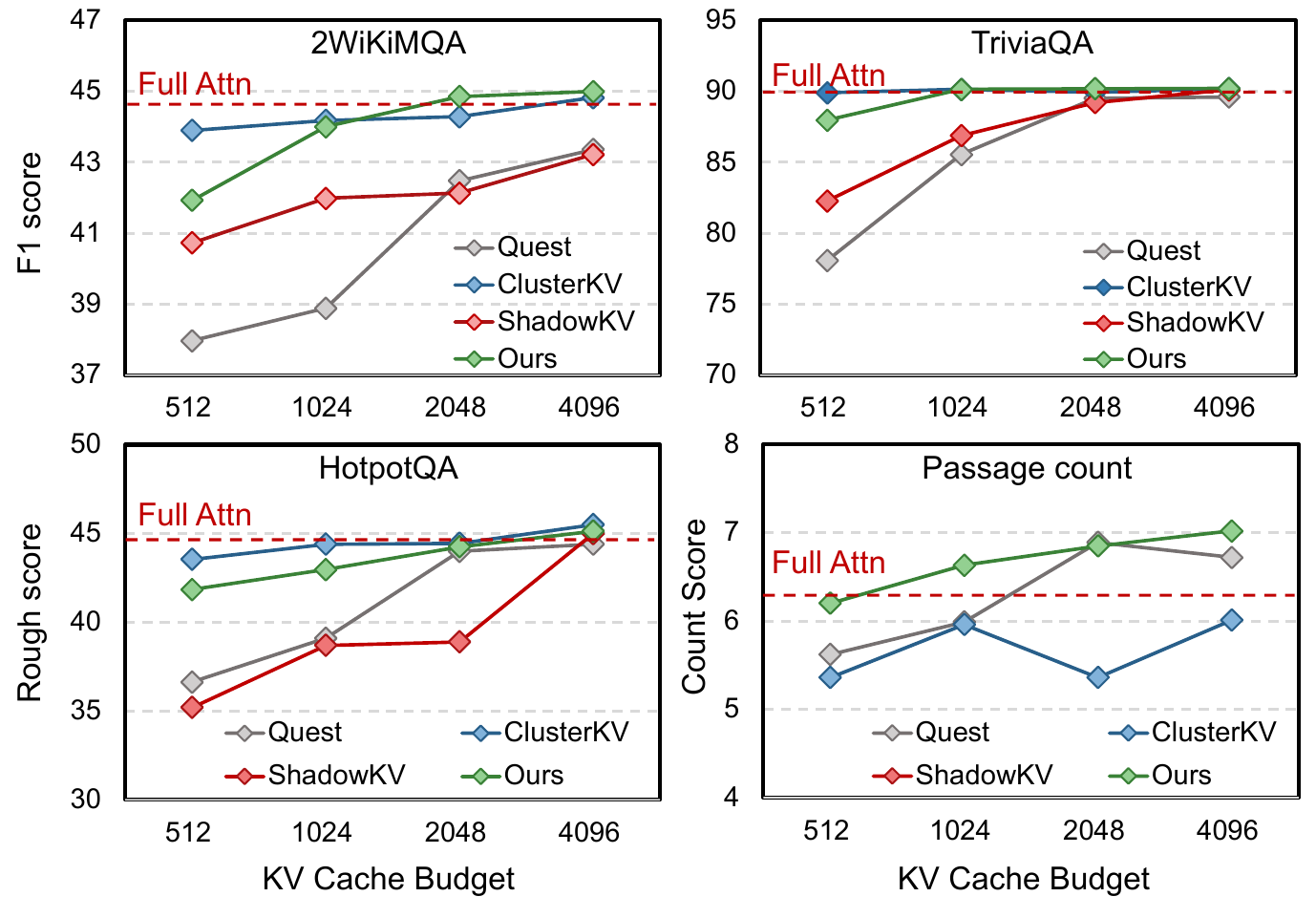}
    \vspace{-20pt}
    \caption{Accuracy  in LongBench on Llama3.1-8B.}
    \vspace{-15pt}
    \label{fig:acc}
\end{figure}

\subsection{Environmental Setup}
We evaluate the performance of \we with various LLMs in two resource-constrained environments, a low-end GPU with limited memory in edge and a high-end GPU with multi-requests in cloud, targeting the long-context input and reasoning scenarios. We compare the performance with several LLM inference engines and some latest works on KV cache optimization in these two environments.

\textbf{Hardware Platforms.}
For the scenario of the high-end GPU with multi-requests in cloud, we choose a workstation with an NVIDIA A100-80GB GPU.
For the scenario of the low-end GPU with limited memory in edge, we select the Lenovo Legion Y7000 PC with NVIDIA RTX 4060 Laptop GPU(8GB) and Intel i7-13650HX CPU. Table~\ref{tab:hardware} shows the detailed hardware specification. 

\begin{table}[!h]
\small
\centering
\caption{Hardware Platforms}
\vspace{-10pt}
\small
\begin{tabular}{@{}c|c|c@{}}
\toprule
                                                     & High-end GPU                                                                                                                    & Low-end GPU                                                                        \\ \midrule
\begin{tabular}[c]{@{}c@{}} GPU\end{tabular} & \begin{tabular}[c]{@{}c@{}} A800, 80GB HBM \\ CUDA 12.1\end{tabular}           & \begin{tabular}[c]{@{}c@{}}RTX 4060 Laptop \\ 8GB GDDR6, CUDA 12.6\end{tabular}        \\ \midrule
CPU                                                  & \begin{tabular}[c]{@{}c@{}}Intel Xeon  Platinum 8358 \\ 1008GB DRAM \end{tabular}   & \begin{tabular}[c]{@{}c@{}} Intel i7-13650HX \\ 24GB DRAM\end{tabular} \\ \bottomrule
\end{tabular}
\label{tab:hardware}
\vspace{-10pt}
\end{table}

\textbf{Baselines.}
To evaluate the performance, we select the typical LLM framework Huggingface~\cite{wolf2020transformers} and the famous LLM inference fast engine, FlashInfer~\cite{yeflashinfer}, as the baselines for full attention.
We also select three latest open-sourced works on KV cache optimization, Quest~\cite{tang2024quest}, ClusterKV~\cite{liu2024clusterkv} and ShadowKV~\cite{sun2024shadowkv} as the baselines for sparse attention.

\textbf{Models and Benchmarks.}
We select three LLMs, Llama3.1-8B~\cite{grattafiori2024llama}, DeepSeek-R1-Distill-Llama-8B~\cite{liu2024deepseek} and Qwen3-8B~\cite{yang2025qwen3} for evaluation in the cloud environment with multiple requests. 
For the edge environment, we select the Reasoning-Llama-3.2-1B~\cite{reasoning-llama3}, which is the reasoning model finetuned on Llama3.2-1B~\cite{grattafiori2024llama}, due to the limited memory. To evaluate the accuracy of \we in two scenarios mentioned above, we select the four tasks(\textit{i.e.}, 2WiKiMQA, TriviaQA, HotpotQA, and Passage count) from LongBench~\cite{bai2024longbench} for the long-context input scenario and the LongWriter benchmark~\cite{bai2024longwriter} for the long-context reasoning scenario.


\begin{figure}[!t]
    \centering
    \includegraphics[width=0.48\textwidth]{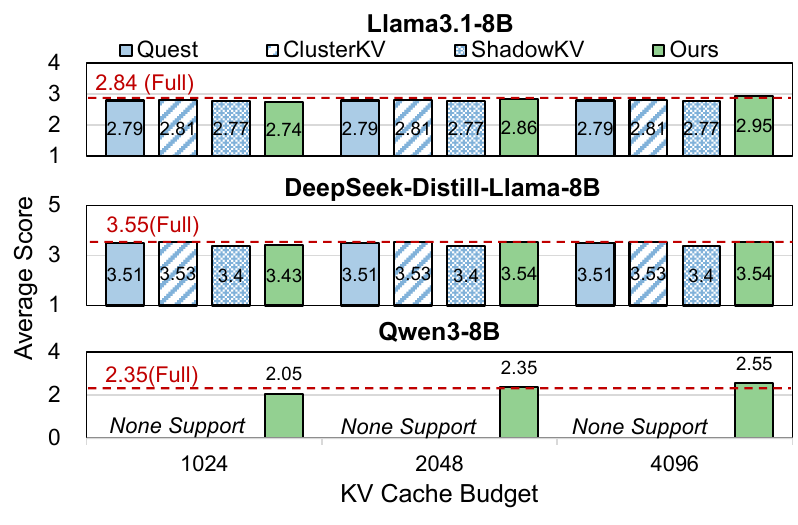}
    \vspace{-20pt}
    \caption{Average score on LongWriter benchmark.}
    \vspace{-15pt}
    \label{fig:long_acc}
\end{figure}

\subsection{Evaluation on Accuracy} \label{sec:eval:acc}
\subsubsection{Results in Long-context Input Scenario.}
Figure~\ref{fig:acc} shows the accuracy results of four tasks in LongBench. Different from existing work, which selects tokens in each layer, we only select globally important tokens before each inference. Therefore, when the budget is small (\textit{i.e.}, 512), our accuracy is slightly lower than ClusterKV~\cite{liu2024clusterkv}. When the budget reaches 1k, \we surpasses the baselines and reaches the accuracy of full attention.

\subsubsection{Results in Long-context Reasoning Scenario.} \label{sec:eval:acc:long_reason}
We use OpenAI GPT-4o to score the output generated from \we and baselines on six dimensions(relevance, accuracy, coherence, clarity, breadth and depth and reading experience). Figure~\ref{fig:long_acc} shows the average scores, and the detailed score is in  Table~\ref{tab:acc_detail} in Appendix~\ref{sec:app:acc}. During experiments, we find that since Quest, ClusterKV and ShadowKV only preprocess the input and retain the KV pair of the newly generated tokens as mentioned in Section~\ref{sec:insight}, the input content, which is only about 100 tokens and less than all the KV budgets, will be completely selected during inference. Therefore, the generated outputs with different KV budgets are the same, resulting in the same scores close to the score of the full attention, but with poor throughput due to the invalid KV optimization.

\begin{table*}[!t]
\small
\centering
\caption{End-to-end throughput (tokens/s) of high-end GPU with multiple requests in cloud. }
\label{tab:multi_req_cloud}
\vspace{-10pt}
\begin{tabular}{@{}ccccccc@{}}
\toprule
Model                                                                                  & {[}In, Out{]} & Full Attn(Eager)                   & Full Attn(Flash Attn)              & Full Attn(FlashInfer)              & ShadowKV                           & Ours                               \\ \midrule
\multirow{4}{*}{\begin{tabular}[c]{@{}c@{}}DeepSeek-Distill\\ -Llama-8B\end{tabular}} & {[}2k, 16k{]} &  45.57({\color[HTML]{9B9B9B}4}, {\color[HTML]{009901}1.00$\times$}) & 145.88({\color[HTML]{9B9B9B}16}, {\color[HTML]{009901}3.20$\times$}) & 490.04({\color[HTML]{9B9B9B}16}, {\color[HTML]{009901}10.75$\times$}) & 366.74({\color[HTML]{9B9B9B}16}, {\color[HTML]{009901}8.05$\times$}) & \uline{\textbf{824.22}}({\color[HTML]{9B9B9B}32}, {\color[HTML]{009901}$18.09\times$}) \\
                                                                                       & {[}2k, 32k{]} &  27.74({\color[HTML]{9B9B9B}4}, {\color[HTML]{009901}$1.00\times$}) & 87.74({\color[HTML]{9B9B9B}8}, {\color[HTML]{009901}$3.16\times$}) & 314.25({\color[HTML]{9B9B9B}8}, {\color[HTML]{009901}$11.32\times$}) & 240.47({\color[HTML]{9B9B9B}16}, {\color[HTML]{009901}$8.67\times$}) & 690.59({\color[HTML]{9B9B9B}32}, {\color[HTML]{009901}\uline{$\textbf{24.89}\times$}}) \\
                                                                                       & {[}16k, 2k{]} &  OOM                                     & 87.71({\color[HTML]{9B9B9B}8}, {\color[HTML]{009901}$1.00\times$}) & 320.41({\color[HTML]{9B9B9B}8}, {\color[HTML]{009901}$3.65\times$}) & 168.06({\color[HTML]{9B9B9B}32}, {\color[HTML]{009901}$1.92\times$}) & 526.47({\color[HTML]{9B9B9B}16}, {\color[HTML]{009901}$6.02\times$}) \\
                                                                                       & {[}32k, 2k{]} &  OOM                                     & 46.89({\color[HTML]{9B9B9B}6}, {\color[HTML]{009901}$1.00\times$}) & 222.06({\color[HTML]{9B9B9B}8}, {\color[HTML]{009901}$4.74\times$}) & 132.07({\color[HTML]{9B9B9B}64}, {\color[HTML]{009901}$2.81\times$}) & 346.88({\color[HTML]{9B9B9B}16}, {\color[HTML]{009901}$7.40\times$}) \\ \midrule
\multirow{4}{*}{Qwen3-8B}                                                             & {[}2k, 16k{]} &  33.77({\color[HTML]{9B9B9B}4}, {\color[HTML]{009901}$1.00\times$}) & 129.67({\color[HTML]{9B9B9B}16}, {\color[HTML]{009901}$3.83\times$}) & 420.12({\color[HTML]{9B9B9B}16}, {\color[HTML]{009901}$12.44\times$}) & - & \uline{\textbf{592.39}}({\color[HTML]{9B9B9B}32}, {\color[HTML]{009901}$17.54\times$}) \\
                                                                                       & {[}2k, 32k{]} &  19.28({\color[HTML]{9B9B9B}4}, {\color[HTML]{009901}$1.00\times$}) & 62.89({\color[HTML]{9B9B9B}8}, {\color[HTML]{009901}$3.26\times$}) & 254.92({\color[HTML]{9B9B9B}8}, {\color[HTML]{009901}$13.22\times$}) & - & 424.92({\color[HTML]{9B9B9B}32}, {\color[HTML]{009901}\uline{$\textbf{22.03}\times$}}) \\
                                                                                       & {[}16k, 2k{]} &  OOM & 60.31({\color[HTML]{9B9B9B}8}, {\color[HTML]{009901}$1.00\times$}) & 259.28({\color[HTML]{9B9B9B}8}, {\color[HTML]{009901}$4.29\times$}) & - & 336.71({\color[HTML]{9B9B9B}16}, {\color[HTML]{009901}$5.58\times$}) \\
                                                                                       & {[}32k, 2k{]} &  OOM & 32.56({\color[HTML]{9B9B9B}6}, {\color[HTML]{009901}$1.00\times$}) & 156.92({\color[HTML]{9B9B9B}6}, {\color[HTML]{009901}$4.81\times$}) & - & 210.75({\color[HTML]{9B9B9B}16}, {\color[HTML]{009901}$6.47\times$}) \\ \bottomrule 
\end{tabular}
\vspace{-10pt}
\end{table*}

\subsection{Evaluation on Speedup and Throughput}
Based on the accuracy evaluation, we select 2048 as the KV budget for the following evaluation. We only select DeepSeek-Distill-Llama-8B and Qwen3-8B for speedup evaluation because Llama3-8B and DeepSeek-Distill-Llama-8B share the same model architecture without impact on speed.
\subsubsection{High-end GPU with multiple requests in cloud.}
\begin{figure}[!t]
    \centering
    \includegraphics[width=0.48\textwidth]{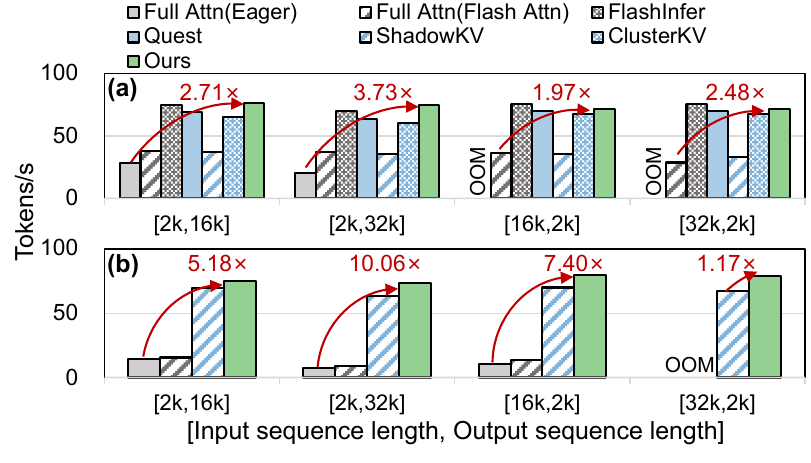}
    \vspace{-20pt}
    \caption{End-to-end throughput with a single request (a) in the cloud environment (b) in the edge environment.}
    \vspace{-10pt}
    \label{fig:single_batch_eval}
\end{figure}

We evaluate \we in two cloud cases, single request and multiple requests, because Quest and ClusterKV only support the single request. Figure~\ref{fig:single_batch_eval}(a) shows the result of a single request case. \we outperforms others in the long-context reasoning scenario because \we effectively reduces the KV cache in attention computation during generation and others use time-consuming preprocessing mentioned in Section~\ref{sec:insight:questions}, but is slightly slower than FlashInfer in long-context input scenario due to the time-consuming retrieval. The results of another case with multiple requests are shown in Table~\ref{tab:multi_req_cloud}. The grey text is the number of requests and the green text is normalized speedup in throughput compared with full attention using Eager implementation in Huggingface. Experiments show that \we achieves up to $24.89\times$ and $2.20\times$ throughput improvement compared with full attention(eager) and state-of-the-art implementation FlashInfer~\cite{yeflashinfer}.

\subsubsection{Low-end GPU with limited memory in edge.}
In the edge scenario, we limit the GPU memory usage to 4GB and compare the performance with full attention(Eagle and Flash Attention~\cite{dao2023flashattention}) and ShadowKV with the offloading strategy. Figure~\ref{fig:single_batch_eval}(b) shows that \we achieves up to $10.06\times$ and $1.17\times$ speedup compared with full attention(eager) and state-of-the-art implementation ShadowKV~\cite{sun2024shadowkv}.

\subsection{Overhead Evaluation}
The overhead in this paper primarily is the memory and training of the retrieval head. As described in Section~\ref{sec:T1}, the retrieval head is obtained through DLM pruning. The weight of the retrieval head for Llama3-8B or Qwen3-8B is only about 60MB. And the K cache is analyzed in Section~\ref{sec:T3}. For the training time, the DLM is provided by EAGLE-3~\cite{li2025eagle-3}, which only needs 24 hours of training using an RTX 3090 GPU for Llama3-8B or Qwen3-8B as described in its paper.

\begin{figure}[!t]
    \centering
    \includegraphics[width=0.48\textwidth]{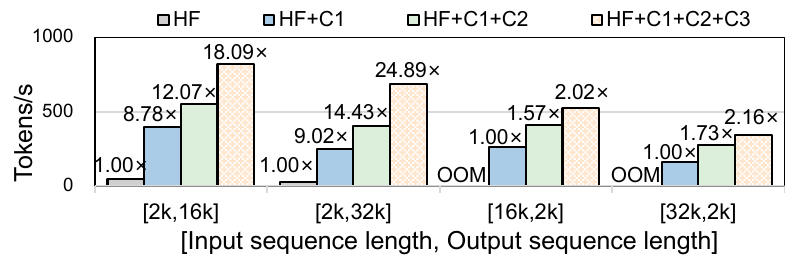}
    \vspace{-20pt}
    \caption{Ablation study of three contributions.}
    \vspace{-10pt}
    \label{fig:ablation}
\end{figure}

\subsection{Ablation Study}
We select the results of the DeepSeek-Distill-Llama-8B in Table~\ref{tab:multi_req_cloud} for the ablation study shown in Figure~\ref{fig:ablation}.

\subsubsection{C1: Lightweight retrieval head}
\we is developed based on the FlashInfer framework~\cite{yeflashinfer}. The speedup of C1 mainly comes from two parts, the FlashInfer framework, which provides a better backend, and sparse attention, which enables more parallel requests processing due to lower memory requirements on the GPU.

\subsubsection{C2: Asynchronous prefetch dataflow}
Due to the heavy KV transfer in C1, the inference is bound to the memory access. Therefore, the speedup improvement of asynchronous dataflow with elastic loading in Figure~\ref{fig:ablation} is mainly from the reduction in data volume.

\subsubsection{C3: Adaptive memory management}
Guided by the theoretical model, we aim to place more KV cache on the GPU to minimize the data transfer overhead, achieving speedup improvement in Figure~\ref{fig:ablation} compared with the HuggingFace with complete offloading.

\section{Conclusion}
In this paper, we analyze the similarity in information focus between the distilled language model(DLM) and the original LLM from
the perspective of information theory, and propose a novel paradigm that leverages a DLM as the retrieval algorithm for important information selection. We think the methodology and perspective can be extended to further studies on machine learning
architecture and system design considering information retrieval.

Based on the paradigm, we present \we,  an algorithm and system co-design for speculative context sparsity in long-context reasoning. \we proposes three contributions at three levels of algorithm, system and compilation for KV retrieval optimization in long-context reasoning, and achieves $24.89\times$ throughput improvement in the cloud environment and $10.06\times$ speedup in the edge environment with negligible accuracy loss, successfully pushing the Pareto frontier of accuracy and speedup.

\begin{acks}
This work was sponsored by Shanghai Rising-Star Program (No. 24QB2706200) and the National Natural Science Foundation of China (No. U21B2031, 62325405).
\end{acks}

\bibliographystyle{ACM-Reference-Format}
\bibliography{ref}

\appendix

\section{LongWriter Benchmark} \label{sec:app:acc}
We provide the detailed score on LongWriter Benchmark in Table~\ref{tab:acc_detail} to support the description in Section~\ref{sec:eval:acc:long_reason}.

\begin{table*}[!h]
\small
\centering
\caption{Detailed Accuracy Results on LongWriter Benchmark}
\label{tab:acc_detail}
\begin{tabular}{@{}l|c|c|c|c|c|c|c|c@{}}
\toprule
Metric & KV Budget &Relevance$\uparrow$ & Accuracy$\uparrow$ & Coherence$\uparrow$ & Clarity$\uparrow$ & Breadth and Depth$\uparrow$ & Reading Experience$\uparrow$ & Average$\uparrow$              \\ \midrule
\multicolumn{8}{l}{\textit{\textbf{Llama3-8B}}}  \\                                                                          
Full & - & 3.73      & 4.07     & 2.21      & 2.64    & 2.625             & 1.77               & 2.84          \\  \midrule
Quest & \multirow{4}{*}{1024}                     & 3.69      & 3.98     & 2.22         & 2.51    & 2.59              & 1.76                  & 2.79      \\
ClusterKV &             & 3.71      & 4.02     & 2.20      & 2.61    & 2.61             & 1.72               & \textbf{2.81}  \\
ShadowKV &  &  3.65     & 3.98    & 2.18   & 2.59    & 2.54 &  1.69     & 2.77 \\
Ours &              & 3.36      & 3.85     & 2.3       & 2.58    & 2.55              & 1.79               & 2.74  \\ \midrule
 
Quest &\multirow{4}{*}{2048}                   & 3.69      & 3.98     & 2.22         & 2.51    & 2.59              & 1.76                  & 2.79      \\
ClusterKV &               & 3.71      & 4.02     & 2.20      & 2.61    & 2.61             & 1.72               & 2.81  \\
ShadowKV &  &  3.65     & 3.98    & 2.18   & 2.59    & 2.54 &  1.69     & 2.77 \\
Ours &              & 3.49      & 3.86     & 2.33      & 2.73    & 2.78              & 1.94               & \textbf{2.86} \\ \midrule
Quest &\multirow{4}{*}{4096}        & 3.69      & 3.98     & 2.22         & 2.51    & 2.59              & 1.76                  & 2.79      \\
ClusterKV &              & 3.71      & 4.02     & 2.20      & 2.61    & 2.61             & 1.72               & 2.81  \\
ShadowKV &  &  3.65     & 3.98    & 2.18   & 2.59    & 2.54 &  1.69     & 2.77 \\

Ours &               & 3.67      & 3.7      & 2.56      & 2.79    & 2.72              & 2.25               & \textbf{2.95} \\ \midrule
\multicolumn{8}{l}{\textit{\textbf{DeepSeek-Distill-Llama-8B}}}  \\                                                                          
Full & - & 4.02      & 4.22     & 3.36      & 3.42    & 3.31             & 3.02               & 3.55          \\  \midrule
Quest & \multirow{4}{*}{1024}                &   3.82    &  3.91     &   3.43    & 3.51    &  3.19             & 3.2               & 3.51          \\
ClusterKV &             & 4.02      & 3.95     & 3.33      & 3.55    & 3.18             & 3.17               & \textbf{3.53}  \\
ShadowKV &  &  3.82     & 3.88    & 3.20   & 3.31    & 3.09 &  3.10    & 3.40 \\
Ours &              & 3.81      & 4.04     & 3.33       & 3.39    & 3.03              & 3.02               & 3.43  \\ \midrule
 
Quest &\multirow{4}{*}{2048}                 &   3.82    &  3.91     &   3.43    & 3.51    &  3.19             & 3.2               & 3.51          \\
ClusterKV &             & 4.02      & 3.95     & 3.33      & 3.55    & 3.18             & 3.17               & 3.53  \\
ShadowKV &  &  3.82     & 3.88    & 3.20   & 3.31    & 3.09 &  3.10    & 3.40 \\
Ours &               & 3.96      & 3.94      & 3.47      & 3.57   & 3.17              & 3.23               & \textbf{3.54} \\ \midrule

Quest &\multirow{4}{*}{4096}                 &   3.82    &  3.91     &   3.43    & 3.51    &  3.19             & 3.2               & 3.51          \\
ClusterKV &             & 4.02      & 3.95     & 3.33      & 3.55    & 3.18             & 3.17               & 3.53  \\
ShadowKV &  &  3.82     & 3.88    & 3.20   & 3.31    & 3.09 &  3.10    & 3.40 \\
Ours &              & 3.96      & 4.02     & 3.45      & 3.55    &3.08               & 3.17              & \textbf{3.54 }      \\ \midrule
\multicolumn{8}{l}{\textit{\textbf{Qwen3-8B}}}  \\                                                                          
Full & - & 2.98      & 3.3     & 1.85      & 2.05    & 2.36             & 1.58               & 2.35          \\ \midrule
\multirow{3}{*}{Ours}  &    1024          & 2.67      & 2.61     & 1.67       & 1.78    & 2.02              &1.52               & 2.05  \\
 
 &2048              & 2.91       & 3.38     & 1.75      & 2.05    &   2.37            & 1.63               &   2.35              \\ 

 &4096               & 3.01      &  3.01    &  2.18       & 2.33    &     2.83         & 1.93                 &  2.55           \\ \bottomrule
\end{tabular}
\end{table*}

\end{document}